\documentclass{article}
\usepackage{ijcai22}

\usepackage{times}

\usepackage{soul}
\usepackage{url}
\usepackage[hidelinks]{hyperref}
\usepackage[utf8]{inputenc}
\usepackage[small]{caption}
\usepackage{graphicx}
\usepackage{amsmath}
\usepackage{booktabs}
\usepackage{bbm}
\usepackage{graphicx}
\usepackage{multirow}
\usepackage{subfigure}
\title{Learning Hierarchical Graph Representation for Image Manipulation Detection}

\author{
Wenyan Pan$^1$\and
Zhili Zhou$^1$\thanks{Corresponding Author}\and
Miaogen Ling$^{1,2}$\footnotemark[1]\and
Xin Geng$^3$\And
Q. M. Jonathan Wu$^4$\\
\affiliations
$^1$School of Computer and Software \& Engineering Research Center of Digital Forensics Ministry of Education, Nanjing University of Information Science and Technology, Nanjing, China\\
$^2$Provincial Key Laboratory for Computer Information Processing Technology, Soochow University, Suzhou, China\\
$^3$School of Computer Science and Engineering, Southeast University, Nanjing, China\\
$^4$Department of Electrical and Computer Engineering, University of Windsor, Windsor, Ontario, Canada
\emails
\{Panwy, mgling\}@nuist.edu.cn,
zhou\_zhili@163.com,
xgeng@seu.edu.cn,
jwu@uwindsor.ca
}

\begin{document}

\maketitle

\begin{abstract}
The objective of image manipulation detection is to identify and locate the manipulated regions in the images. Recent approaches mostly adopt the sophisticated Convolutional Neural Networks (CNNs) to capture the tampering artifacts left in the images to locate the manipulated regions. However, these approaches ignore the feature correlations, i.e., feature inconsistencies, between manipulated regions and non-manipulated regions, leading to inferior detection performance. To address this issue, we propose a hierarchical Graph Convolutional Network (HGCN-Net), which consists of two parallel branches: the backbone network branch and the hierarchical graph representation learning (HGRL) branch for image manipulation detection. Specifically, the feature maps of a given image are extracted by the backbone network branch, and then the feature correlations within the feature maps are modeled as a set of fully-connected graphs for learning the hierarchical graph representation by the HGRL branch. The learned hierarchical graph representation can sufficiently capture the feature correlations across different scales, and thus it provides high discriminability for distinguishing manipulated and non-manipulated regions. Extensive experiments on four public datasets demonstrate that the proposed HGCN-Net not only provides promising detection accuracy, but also achieves strong robustness under a variety of common image attacks in the task of image manipulation detection, compared to the state-of-the-arts.

\end{abstract}

\begin{figure}[h]
\centering
\includegraphics[width=7cm,height=5.5cm]{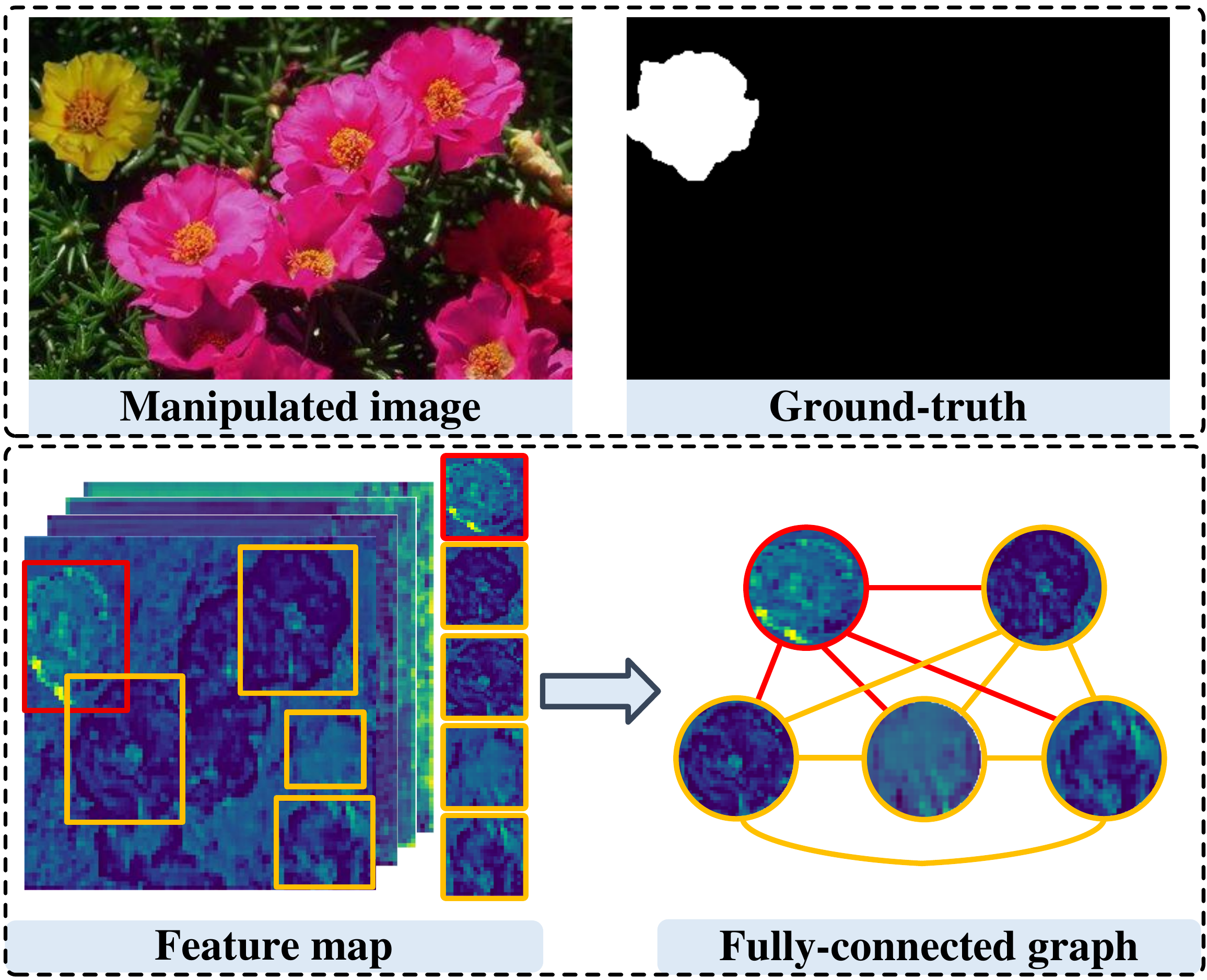}
\caption{Illustration of the importance of feature correlations in manipulation detection. The feature map of the manipulated and non-manipulated regions are shown in the red box and yellow box, respectively. In the bottom left figure, the feature map of the manipulated region is quite distinct and inconsistent with those of the non-manipulated regions, while the feature maps of non-manipulated regions are not. In the bottom-right figure, the feature correlations within the feature map are modeled as a fully-connecting graph. The contrast of the feature map is enhanced for better visualization.}
\label{Figure:1}
\end{figure}

\section{Introduction}

Recently, determination of image authenticity has become a challenging problem. Due to the availability of powerful image-editing tools, the malicious users can manipulate digital images easily without leaving obvious manipulation traces. The existing image manipulation operations usually modify the image content by adopting several operations, i.e., copy-move, splicing, and removal. Thus, as an effective way to determine image authenticity, image manipulation detection has received more and more attention.

The basic idea of image manipulation detection is to capture the tampering artifacts to identify and localize the manipulated regions for image authenticity determination. The traditional manipulation detection approaches mainly extract handcrafted features to identify and localize manipulated regions, such as Color Filter Array (CFA) ~\cite{popescu2005exposing}, Discrete Cosine Transform (DCT) ~\cite{li2017image}, Scale-Invariant Feature Transform (SIFT) ~\cite{zhou2016effective}. Such approaches have achieved impressive performance. However, since the feature extraction algorithms are designed manually, these approaches tend to locate manipulated regions in only one or certain specific types of manipulations.

To address the generalization issue of the traditional approaches, some recent works ~\cite{bappy2017exploiting,wu2018busternet} exploited different CNN architectures to learn manipulation features automatically for image manipulation detection. Some other approaches ~\cite{bayar2018constrained,zhou2018learning,yang2020constrained,yang2020constrained,wu2019mantra} utilized some high-pass filters or novel CNN layers to suppress image content and adaptively learn more implicit manipulation traces for image manipulation detection.

Even though CNNs-based approaches have shown impressive performance in image manipulation detection, they ignore the feature correlations between manipulated and non-manipulated regions, which play an important role in manipulation detection. As a result, the detection accuracy is compromised significantly. To further explain the importance of the feature correlations in manipulation detection, we visualize a feature map of a given image in Figure~\ref{Figure:1}. As shown in Figure~\ref{Figure:1}, in a feature map of the image, the manipulated region and non-manipulated regions are shown by the red box and yellow box, respectively. It can be clearly observed that the feature map of the manipulated region is quite distinct and inconsistent with those of the non-manipulated regions, while the feature maps of non-manipulated regions are not. Thus, the feature correlations, i.e., feature inconsistency, between manipulated and non-manipulated regions will provide important clues for distinguishing the two kinds of regions.

Motivated by the above observation, this paper proposes a novel network architecture for image manipulation detection, called HGCN-Net. The proposed HGCN-Net consists of two parallel branches: the backbone network branch and the hierarchical graph representation learning (HGRL) branch. First, the backbone network branch extracts the feature maps from a given image. Second, the HGRL branch constructs a set of fully-connected graphs by connecting all grids in each of the feature maps selected from different convolutional layers, and learns the hierarchical graph representation from those fully-connected graphs. Finally, the learned hierarchical graph representation is transformed to the original CNN space and then is fused with feature maps in the backbone network branch for localizing the manipulated regions in pixel-level.

The main contributions of this paper are as follows:
\begin{itemize}
\item The feature correlations within the selected feature maps are modeled as a set of fully-connected graphs. To the best of our knowledge, this is the first network that explores the graph structure for image manipulation detection.
\item The HGCN-Net is proposed to learn the hierarchical graph representation from those fully-connected graphs. The learned representation can sufficiently capture the feature correlations across different scales, and thus it provides high discriminability for distinguishing manipulated and non-manipulated regions. In practice, the manipulated images transmitted on the Internet usually suffer from a variety of intentional or unintentional attacks such as JPEG compression and Gaussian noise addition. Since the feature correlations are relatively stable under those attacks, the learned representation also shows strong robustness and provides desirable results of manipulation detection under those attacks.
\item Extensive experiments on four public manipulated image datasets demonstrate that the proposed HGCN-Net achieves promising detection performance and strong robustness under a variety of common image attacks, compared to the state-of-the-arts.
\end{itemize}

\begin{figure*}[htbp]
\centering
\includegraphics[width=11cm,height=5cm]{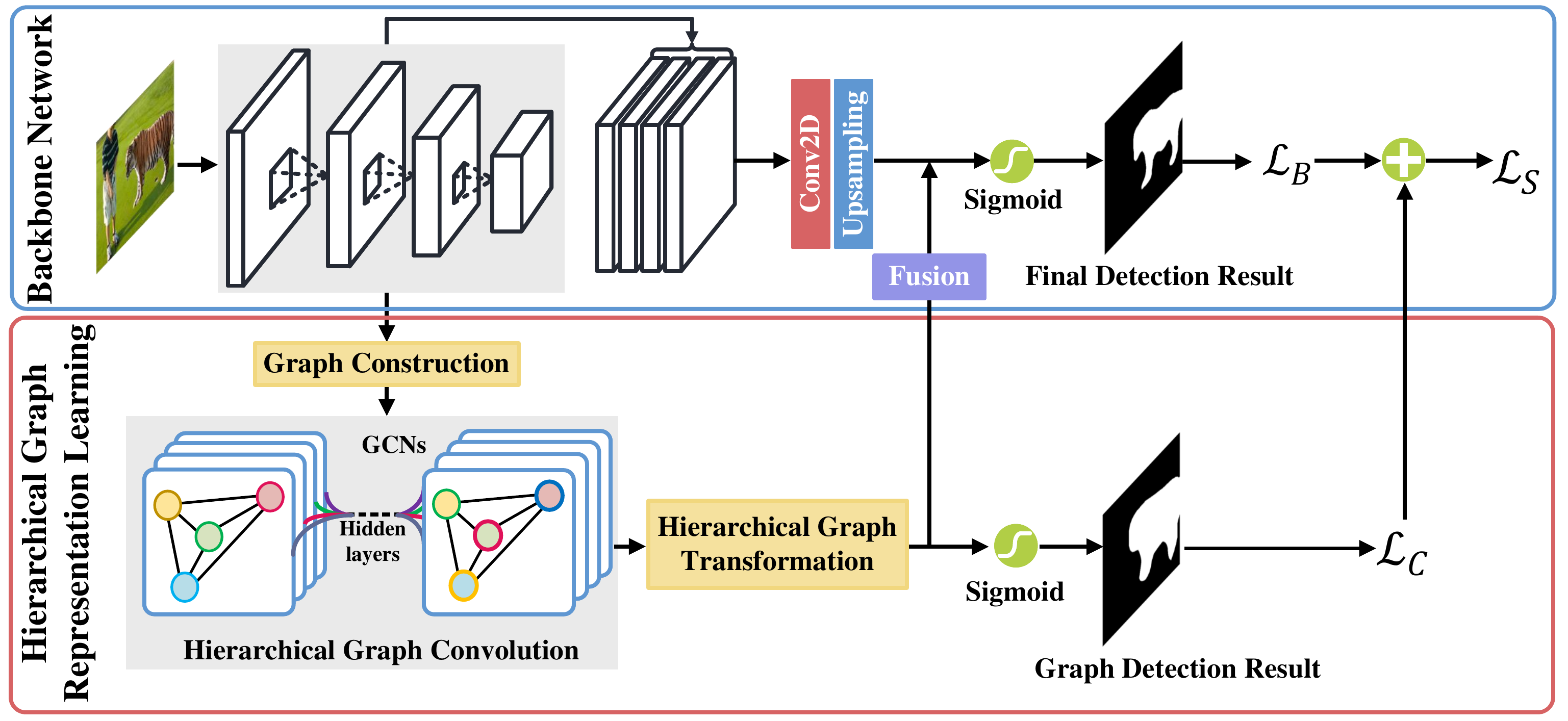}
\caption{The pipeline of the HGCN-Net for image manipulation detection.}
\label{Figure:2}
\end{figure*}

\section{Related Works}
\subsection{Image Manipulation Detection}

{\bf Handcrafted feature-based approaches.} In early work, the image manipulation detection approaches usually extract handcrafted features, such as Discrete Cosine Transform (DCT) ~\cite{li2017image}, Scale-Invariant Feature Transform (SIFT) ~\cite{zhou2016effective}, to capture the tampering traces left in the images for localizing the manipulated regions. However, since the handcrafted features can only capture the traces caused by one or several certain tampering operations, leading to limited generalization ability of these approaches.

\noindent {\bf CNN-based approaches.} To improve the generalization ability of hancrafted feature-based approaches, some approaches ~\cite{salloum2018image,wu2018busternet,zhou2020generate} explore the sophisticated CNNs to automatically learn the tampering trace features for image manipulation detection. Some other approaches suppress the image content to facilitate the extraction of tampering trace features. ~\cite{zhou2018learning} proposed a two-stream Faster R-CNN, which pays more attention to the noise feature to locate the manipulated regions. ~\cite{rao2016deep} used the Steganalysis Rich Model (SRM) ~\cite{fridrich2012rich} to initialize the kernel of the first convolution layer with steganalysis filter kernels, aiming to suppress the effects of image content for manipulation detection. ~\cite{bayar2018constrained} proposed the constrained convolutional layer to adaptively learn the tampering trace features. ~\cite{yang2020constrained} proposed a two-stream network, which combines Convolutional Block Attention Module (CBAM) ~\cite{woo2018cbam} and SRM to learn the tampering trace features for image manipulation detection. Although these approaches generally outperform the handcrafted feature-based approaches, they do not take into account the feature correlations between manipulated and non-manipulated regions. Thus, there is still a large room for performance improvement.

\subsection{Graph Convolutional Network}
The Graph Convolutional Networks (GCNs)~\cite{kipf2016semi} and their variants have been applied in many computer vision tasks, such as image classification ~\cite{knyazev2019image}, object detection ~\cite{zhao2021graphfpn}, segmentation ~\cite{lu2020cnn,li2018beyond}. ~\cite{knyazev2019image} built multi-graphs by segmenting each image as a set of superpixels and formulated image classification as a multigraph classification problem. ~\cite{9170817} investigated the CNNs and miniGCN for hyperspectral image classification. In ~\cite{zhao2021graphfpn}, the graph feature pyramid network is proposed to capture the intrinsic image structures across different scales by mapping the superpixels of each image to the graph nodes. ~\cite{li2018beyond} formulated the semantic segmentation as the graph node classification task for semantic segmentation. Inspired by these GCN models, we attempt to learn the graph representation from images to describe the feature correlations between manipulated and non-manipulated regions for image manipulation detection.

\section{The Proposed Approach}

\subsection{Pipeline of HGCN-Net}
The pipeline of the proposed HGCN-Net is illustrated in Figure~\ref{Figure:2}. The proposed HGCN-Net consists of two branches: the backbone network branch and the hierarchical graph representation learning (HGRL) branch. First, the backbone network branch extracts the feature maps for the graph construction in the HGRL branch; Second, the HGRL branch is implemented by three following steps: constructing graphs from the feature maps, learning the hierarchical graph representation from the constructed graphs by hierarchical graph convolution, and transforming the learned representation to the original CNN space. Finally, the backbone network branch integrates the feature maps with the hierarchical graph representation learned by the HGRL branch to obtain the final detection results. The main contribution of HGCN-Net lies in the three steps of the HGRL branch. Therefore, in the following sections, we will elaborate each of the three steps, and introduce the loss function used in HGCN-Net.

\subsection{Graph Construction}

The graph construction is illustrated by the green box of Figure~\ref{Figure:3}. After extracting the selected multi-scale feature maps from the different layers in the backbone network branch, we construct a set of fully-connected graphs to model the feature correlations by connecting all grids in each feature map. The multi-scale feature maps represent as $\mathcal{P}=\left\{\textbf{\emph{P}}_1, \textbf{\emph{P}}_2, \textbf{\emph{P}}_3, \textbf{\emph{P}}_4 \right\}$, where $\textbf{\emph{P}}_i \in \mathbbm{R}^{B \times C_i \times W_i \times H_i}$, $i=1,2,...,4$, $B$, $C$, $W$, and $H$ denotes the batch size, channel, width, and height of the feature map, respectively. Note that there is a large amount of grid in each $\textbf{\emph{P}}_i$, and thus the construction of those fully-connected graphs is time-consuming. To reduce time consumption, we first transform $\mathcal{P}$ into a low-dimensional feature maps $\mathcal{P^{'}}=\left\{\textbf{\emph{P}}_1^{'}, \textbf{\emph{P}}_2^{'}, \textbf{\emph{P}}_3^{'}, \textbf{\emph{P}}_4^{'} \right\}$ by down-sampling process $down_\downarrow$. Formally, $\mathcal{P^{'}}$ is computed as:

\begin{equation}
    \mathcal{P^{'}}=down_\downarrow(\mathcal{P}),\quad \textbf{\emph{P}}_i^{'} \in \mathbbm{R}^{B \times C_i \times W_i^{'} \times H_i^{'}}
\end{equation}

\noindent where $W_i^{'}=W_i/s_i$, $H_i^{'}=H_i/s_i$, $s_i=\left\{ s_1, s_2, s_3, s_4 \right\}$ denotes the down-sampling factor. The down-sampling process is achieved by implementing a convolution operation with $3\times3$ kernel and stride of $2$, and batch normalization operation.

After the down-sampling process, the set of fully-connected graphs $\mathcal{G}=\left\{ \mathcal{G}_1, \mathcal{G}_2, \mathcal{G}_3, \mathcal{G}_4 \right\}$ is constructed by connecting all grids in each $\textbf{\emph{P}}_i^{'}$. The $i$-th fully-connected graph is represented as $\mathcal{G}_i=\left( \mathcal{V}_i, \mathcal{E}_i \right)$, where $\mathcal{V}_i$ denotes a set of $N_i$ graph nodes, $N_i=W_i^{'}\times H_i^{'}$, and $\mathcal{E}_i$ are set of edges represented as an adjacency matrix $A_i \in \mathbbm{R}^{N_i \times N_i}$, representing the spatial distances between every two graph nodes.

\subsection{Hierarchical Graph Convolution}

\begin{figure}[t]
\centering
\includegraphics[width=8.5cm,height=2cm]{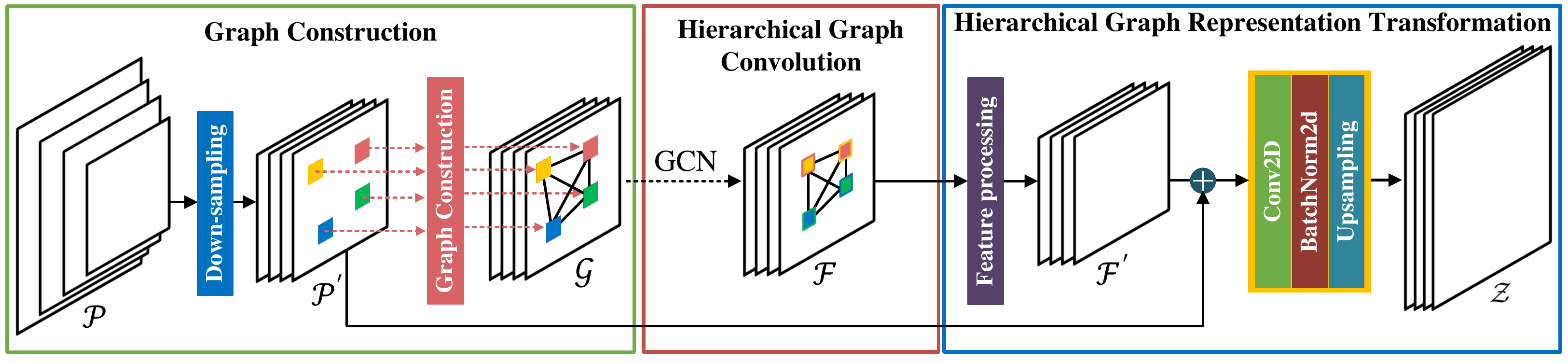}
\caption{The illustration of the HGRL branch, which aims to implement the three steps: graph construction, hierarchical graph convolution, and hierarchical graph representation transformation.}
\label{Figure:3}
\end{figure}

After the graph construction, the GCN is introduced to perform convolution operations on $\mathcal{G}$ to learn the hierarchical graph representation. The red box of Figure~\ref{Figure:3} illustrates the hierarchical graph convolution. The convolution operation takes the $\mathcal{G}$ and corresponding feature matrix $\textbf{\emph{P}}_i^{'}$ as input. After implementing a 2-layer graph convolution operation on $\mathcal{G}$ and $\textbf{\emph{P}}_i^{'}$, we obtain the hierarchical graph representation $\mathcal{F}=\left\{\textbf{\emph{F}}_1, \textbf{\emph{F}}_2, \textbf{\emph{F}}_3, \textbf{\emph{F}}_4 \right\}$ by

\begin{equation}
   \emph{\textbf{F}}_i=\sigma_s\left( \widehat{\textbf{\emph{A}}}_i \sigma_r  \left( \widehat{\textbf{\emph{A}}}_i \textbf{\emph{P}}_i^{'} \textbf{\emph{M}}_i^{\left(0 \right)} \right )\textbf{\emph{M}}_i^{1}\right)
\end{equation}

\noindent where ${\textbf{\emph{F}}}_i \in \mathbbm{R}^{B \times C_i \times N^2_i}$, $\textbf{\emph{A}}_i$ denotes the adjacency matrix. $\textbf{\emph{M}}_i^{\left(0 \right)}$ and $\textbf{\emph{M}}_i^{\left(1 \right)}$ denote the weight matrix for GCN in the first and second layer. $\sigma_s$ and $\sigma_r$ denote the softmax and ReLU activation function, respectively. $\widehat{\textbf{\emph{A}}}_i$ denotes the preprocessing step, represented by

\begin{equation}
    \widehat{\textbf{\emph{A}}}_i= \widetilde{\textbf{\emph{D}}_i}^{-1/2} \widetilde{\textbf{\emph{A}}_i} \widetilde{\textbf{\emph{D}}_i}^{-1/2}
\end{equation}

\noindent where $ \widehat{\textbf{\emph{A}}}_i=\textbf{\emph{A}}_i+\textbf{\emph{I}}$ denotes an adjacency matrix with self-loops, $\textbf{\emph{I}}$ denotes an identity matrix. $\widetilde{\textbf{\emph{D}}_i}$ denotes is the diagonal node degree matrix of $\widetilde{\textbf{\emph{A}}_i}$. Then, the learned hierarchical graph representation $\mathcal{F}$ will be further transformed to the CNN space in the following step.

\subsection{Hierarchical Graph Representation Transformation}

To enhance the discriminability of the CNN features, the learned hierarchical graph representation will be fused with the CNN features in the backbone network branch. However, the GCN and CNN are capable of learning distinctive representations of manipulated images in different aspects, and the form of the learned hierarchical graph representation is quite different from that of CNN representations. To make the forms of the two representations consistent, the learned hierarchical graph representation $\mathcal{F}$ is transformed to the original CNN space at first. The blue box of Figure~\ref{Figure:3} illustrates the step of hierarchical graph representation transformation. After obtaining the hierarchical graph representation $\mathcal{F}$, we map the $\mathcal{F}$ back into the original CNN feature map. Specifically, we first transform $\mathcal{F}$ into $\mathcal{F}^{'}=\left\{\textbf{\emph{F}}_1^{'}, \textbf{\emph{F}}_2^{'}, \textbf{\emph{F}}_3^{'}, \textbf{\emph{F}}_4^{'} \right\}$ by reshaping, where $\textbf{\emph{F}}_i^{'} \in \mathbbm{R}^{B \times C_i \times W_i^{'} \times H_i^{'}}$. Moreover, inspired by skip connection, we integrate the $\mathcal{F}^{'}$ with feature maps $\mathcal{P}^{'}$ by element-wise addition, the convolution operation and batch normalization operation. Finally, the bilinear interpolation is applied to generate the final hierarchical graph representation $\mathcal{Z}=\left\{\textbf{\emph{Z}}_1, \textbf{\emph{Z}}_2, \textbf{\emph{Z}}_3, \textbf{\emph{Z}}_4 \right\}$ by

\begin{equation}
    {\textbf{\emph{Z}}}_i= up\left(BN\left( Conv1\left( \textbf{\emph{F}}_i^{'}\oplus \textbf{\emph{P}}_i^{'} \right) \right) \right)
\end{equation}

\noindent where the $up	\left( \cdot \right)$ denotes upsampling with bilinear interpolation. $\oplus$ denotes the element-wise addition. $Conv1\left(\cdot\right)$ denotes a convolution operation with $1\times1$ kernel and stride of 2. $BN\left(\cdot\right)$ denotes a batch normalization operation.

After the step of hierarchical graph representation transformation, $\mathcal{Z}$ is fused with the output feature maps of the upsampling in the backbone network branch for further detection. Two fusion operations are adopted. First, $\mathcal{Z}$ is fused with the output feature maps of the upsampling in the backbone network branch by element-wise addition. Second, $\mathcal{Z}$ is fused with the output feature maps of the upsampling in the backbone network branch by concatenating and convolving with a standard convolution layer. After that, the fused representation is processed by the sigmoid function to generate the final result.

\subsection{Loss Function}

In the training stage of the proposed HGCN-Net, we design the principal loss function $\mathcal{L}_S$ to supervise the detection results of the HGCN-Net. The designed loss function contains two parts. The loss $\mathcal{L}_C$ is utilized to supervise the detection results of the HGRL branch for optimizing the hierarchical graph representation for further fusion. While the loss $\mathcal{L}_B$ is utilized to supervise the detection results of the backbone network branch and the performance of the fusion operation. All the loss functions are Dice Loss. Furthermore, we use the parameter $\alpha$ to control the importance of the $\mathcal{L}_C$ and $\mathcal{L}_B$ during training. The principal loss function $\mathcal{L}_S$ is represented by
\begin{equation}
\resizebox{.91\linewidth}{!}{$
    \displaystyle
    \mathcal{L}_S\left(X_m,g_t\right) = \alpha \mathcal{L}_C \left(R\left(X_m\right),g_t\right) + \left(1-\alpha\right) \mathcal{L}_B \left(X_m,g_t\right)
$}
\end{equation}%

\noindent where $X_m$ denote an input manipulated image, $R\left(X_m \right)$ denotes the detection results of the HGRL branch.

\section{Experiments}

\subsection{Datasets and Evaluation Metrics}

To validate the effectiveness of the proposed HGCN-Net, we conduct extensive experiments on the four most widely used datasets: CASIA ~\cite{dong2013casia}, COVER ~\cite{wen2016coverage}, NIST16 ~\cite{guan2019mfc}, and Columbia ~\cite{hsu2006columbia}. The CASIA consists of two parts: CASIA 1.0 and CASIA 2.0, which contain 921 and 5124 manipulated images, respectively. We choose CASIA 2.0 for training and CASIA 1.0 for testing. The COVER dataset contains 100 manipulated images, in which we choose 75 images for training and 25 images for testing. The NIST16 contains 564 manipulated images, in which we choose 404 images for training and 160 for testing. The Columbia dataset, consisting of 180 manipulated images, is split in half to construct the train/test sets.

Similar to ~\cite{zhou2020generate,rao2016deep}, we use pixel-level $F_1$ score and Matthews Correlation Coefficient (MCC) for evaluating performance of the proposed HGCN-Net on the four benchmarks, we set the prediction threshold as 0.5 to get a binary prediction mask.

\subsection{Implementation Details}

The HGCN-Net is implemented on the Pytorch with an NVIDIA RTX 3090 GPU. We apply the Adam optimizer with batch size 64 for CASIA and 8 for other datasets. The initial learning rate is 0.0001 with decaying 6\% every two epochs. Training is terminated after 100 epochs. During the training, we use data augmentation techniques including flipping, rotation, padding, cropping, and adding Gaussian noise using open-source toolbox albumentation. We use the pretrained model on ImageNet to initialize the backbone network branch in HGCN-Net to improve the training stability.

During the training process, the input images are resized to $320 \times 320$, $480\times480$, $512\times512$,and $512\times512$ for CAISA, COVER, NIST16, and Columbia, respectively. For the hierarchical graph projection, we set three different down-sampling factors $s_i$, e.g., $s_{1i}=\left\{16, 8, 4, 2 \right\}$,
$s_{2i}=\left\{8, 4, 2, 1 \right\}$, and $s_{3i}=\left\{4, 2, 1, 1 \right\}$, note that $\textbf{\emph{P}}_4$ does not need a down-sampling operation since the size of the feature map is small enough. For the loss function, For the loss function, we set the weight $\alpha$ as different values, i.e., $\alpha=\left\{0.1, 0.3, 0.5, 0.7,0.9 \right\}$ for comparison.

\begin{table}[t]
\centering
\begin{tabular}{lll}
\hline
Approach  & $ F_1 (\%) $ & MCC (\%) \\
\hline
R-101-FPN       & 38.29  & 37.85     \\
R-101-FPN+GR($\textbf{\emph{Z}}_1$)-A       & 38.79  & 37.99      \\
R-101-FPN+GR($\textbf{\emph{Z}}_2$)-A    & 38.79  & 38.16     \\
R-101-FPN+GR($\textbf{\emph{Z}}_3$)-A   & 39.56  & 38.89     \\
R-101-FPN+GR($\textbf{\emph{Z}}_4$)-A   & 39.70  & 38.96     \\
R-101-FPN+HGR-A   & 39.71  & 39.30     \\
R-101-FPN+HGR-A(w/ $\mathcal{L}_S$)   & 40.06  & 39.19     \\
R-101-FPN+HGR-C(w/ $\mathcal{L}_S$)   & \textbf{40.81}  & \textbf{39.39}     \\
\hline
\end{tabular}
\caption{Ablation study of HGCN-Net on CASIA with different settings.}
\label{tab:1}
\end{table}

\begin{table}[t]
\centering
\begin{tabular}{llll}
\hline
Approach  & $ F_1 (\%) $ & MCC (\%) & Speed \\
\hline
HGCN-Net ($s_{1i}$)       & 40.13  & 38.64  & \textbf{497.23}   \\
HGCN-Net ($s_{2i}$)      & \textbf{40.81}  & \textbf{39.39}  & 682.98    \\
HGCN-Net ($s_{3i}$)    & 39.10  & 38.36 & 3613.52    \\
\hline
\end{tabular}
\caption{Ablation study of HGCN-Net on CASIA with different down-sampling factors.}
\label{tab:2}
\end{table}

\begin{table}[t]
\centering
\begin{tabular}{lll}
\hline
Approach  & $ F1 (\%) $ & MCC (\%) \\
\hline
HGCN-Net ($\alpha=0.1$)       & 39.17  & 38.41     \\
HGCN-Net ($\alpha=0.3$)      & 39.11  & 37.17      \\
HGCN-Net ($\alpha=0.5$)    & \textbf{40.81}  & \textbf{39.39}     \\
HGCN-Net ($\alpha=0.7$)    & 38.88  & 38.24     \\
HGCN-Net ($\alpha=0.9$)    & 39.53  & 38.83     \\
\hline
\end{tabular}
\caption{Ablation study of HGCN-Net on CASIA with different weights of the loss function.}
\label{tab:3}
\end{table}

\subsection{Experimental Results}

{\bf Ablation Study of network architecture.} To validate the impact of different network structures, we conduct ablation studies incrementally on CASIA. We test and compare the HGCN-Net with different settings: (1) R-101-FPN, in which ResNet101 network with the Feature Pyramid Network is used as the backbone network. (2) R-101-FPN+GR$(\textbf{\emph{Z}}_i)$-A, in which R-101-FPN separately fuses with each graph representation $\textbf{\emph{Z}}_1$, $\textbf{\emph{Z}}_2$, $\textbf{\emph{Z}}_3$, and $\textbf{\emph{Z}}_4$ by element-wise addition. (3) R-101-FPN+HGR-A(w/ $\mathcal{L}_B$), in which R-101-FPN fuses with the hierarchical graph representation $\mathcal{Z}$ by element-wise addition. (4) R-101-FPN+HGR-A(w/ $\mathcal{L}_S$), in which R-101-FPN fuses with the hierarchical graph representation $\mathcal{Z}$ by element-wise addition with the designed loss function $\mathcal{L}_S$. (5) R-101-FPN+HGR-C(w/ $\mathcal{L}_S$), in which R-101-FPN fuses with the hierarchical graph representation $\mathcal{Z}$ by concatenation operation with the designed loss function $\mathcal{L}_S$.

As shown in Table~\ref{tab:1}, R-101-FPN achieves 38.29\% in $F_1$ score and 37.85\% in MCC. R-101-FPN+GR($\textbf{\emph{Z}}_i$)-A generally outperforms R-101-FPN in $F_1$ score and MCC. Meanwhile, when we adopt R-101-FPN+HGR-A, the network improves the $F_1$ score and MCC performance over the R-101-FPN by 1.42\% and 1.45\%. Furthermore, when we adopt R-101-FPN+HGR-A(w/ $\mathcal{L}_S$), we achieve higher $F_1$ score and MCC, i.e., 40.06\% and 39.19\%, when using the designed loss function $\mathcal{L}_S$. Finally, it is clear that R-101-FPN+HGR-C(w/ $\mathcal{L}_S$) outperforms the R-101-FPN by 2.52\% in $F_1$ and 1.54\% in MCC, and achieves highest accuracy, i.e., 40.81\% in F1 score and 39.39\% in MCC, which indicate the concatenation operation further improve the performance. That might be because the concatenation fusion operation retains more graph representation, which is beneficial for detection accuracy. According to the above, we choose R-101-FPN+HGR-C(w/ $\mathcal{L}_S$) as the default setting of HGCN-Net in the following experiments.

{\bf Ablation Study of different down-sampling factors.} We also perform an ablation study of different down-sampling factors $s_i$. The different down-sampling factors $s_i$ lead to different sizes of the graphs. We also test the speed (inference time) of these approaches. Speed is measured by calculating the average inference time of the testing set on CASIA. From the results of Table~\ref{tab:2}, it can be clearly observed that $s_{2i}=\left\{8, 4, 2, 1 \right\}$ yields the best performance with acceptable speed. So, we choose the down-sampling factor $s_{2i}$ as the default setting in all the following experiments.

\begin{figure}[t]
\centering
\includegraphics[width=8.7cm,height=5.8cm]{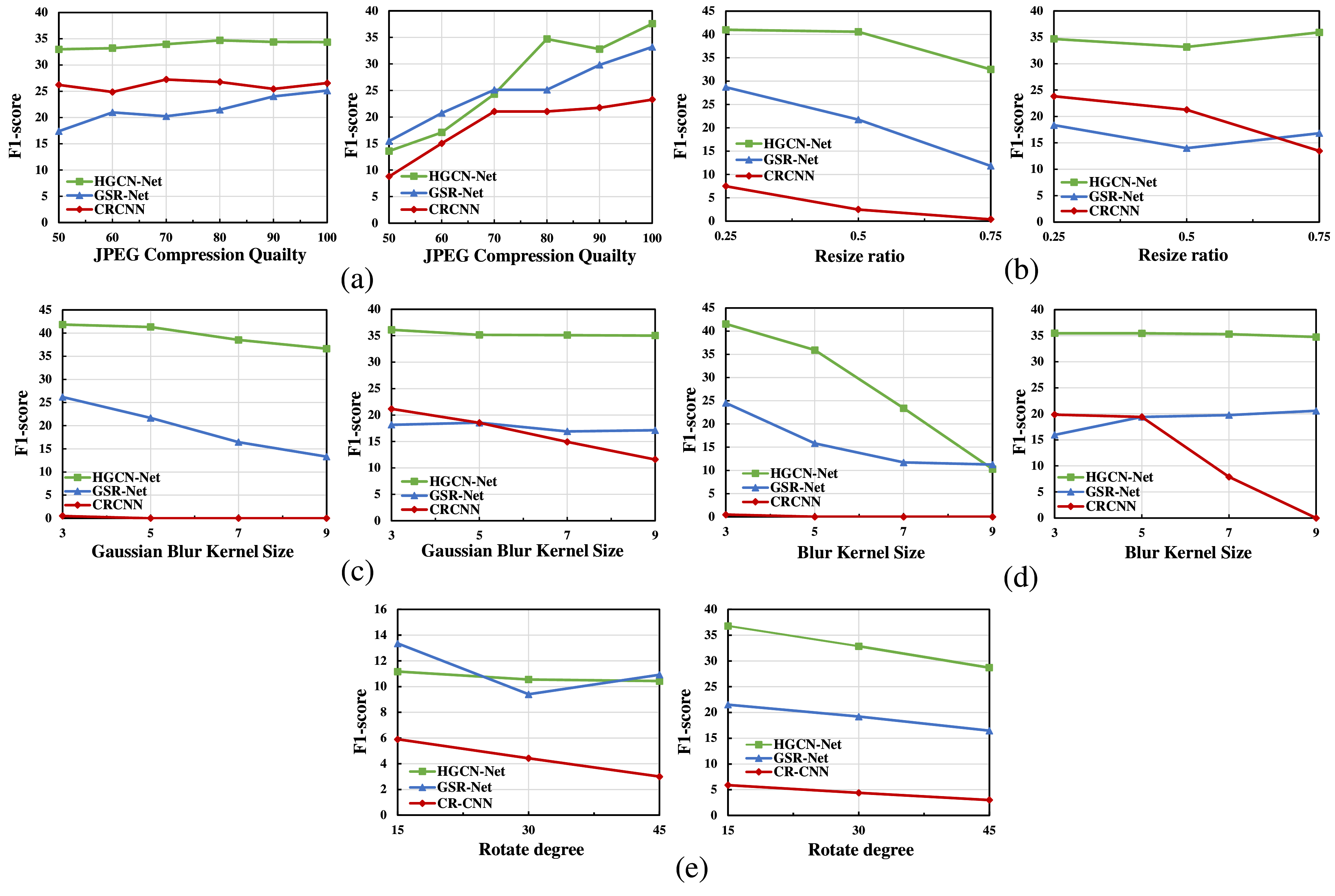}
\caption{Robustness comparison between different image attacks of proposed HGCN-Net on CASIA and COVER datasets.}
\label{Figure:4}
\end{figure}

{\bf Ablation Study of the weight of the loss function.} In this part, we design an ablation study for $\alpha$, we test the different values of $\alpha=\left\{0.1, 0.3, 0.5, 0.7,0.9 \right\}$. From the results of Table~\ref{tab:3}, it can be clearly observed that $\alpha=0.5$ yields the best performance. Thus, in all following experiments, we choose the $\alpha=0.5$ as the default setting.

\begin{table*}[t]
\centering
\begin{tabular}{lllllllll}
\hline
\multirow{2}{*}{Approach} & \multicolumn{2}{l}{CASIA} & \multicolumn{2}{l}{NIST16} & \multicolumn{2}{l}{COVER} & \multicolumn{2}{l}{Columbia} \\ \cline{2-9}
                  &          $F_1$ (\%) & MCC (\%)     & $F_1$ (\%) & MCC (\%)  & $F_1$ (\%) & MCC (\%) & $F_1$ (\%) & MCC (\%)   \\ \hline
                ResNet101+FPN  &   38.29 & 37.85	&	66.26 & 65.86 & 36.85 & 30.65	&91.80&	89.41\\
                EXIF-Consistency  &   - & -	&	- & - & - & -	&88.00&	80.00    \\
                CR-CNN  &   24.68	 & 26.12	& \textbf{89.87} &	\textbf{89.59} &	26.64&	19.90&	91.18	&88.07       \\
                GSR-Net  &    35.80	 & 35.97	& - &	- &	25.27&	23.34&	-	&-        \\
                HGCN-Net  & \textbf{40.81}	 & \textbf{39.39}	& 71.02 &	70.79 &	\textbf{39.32}&	\textbf{36.94}&	\textbf{91.81}	&\textbf{89.76}
\\ \hline
\end{tabular}
\caption{$F_1$ score and MCC comparison of different approaches on four public datasets. $``-"$ denotes that the results are not available in the references}
\label{tab:4}
\end{table*}

\begin{figure}[t]
\centering
\includegraphics[width=8.7cm,height=6.2cm]{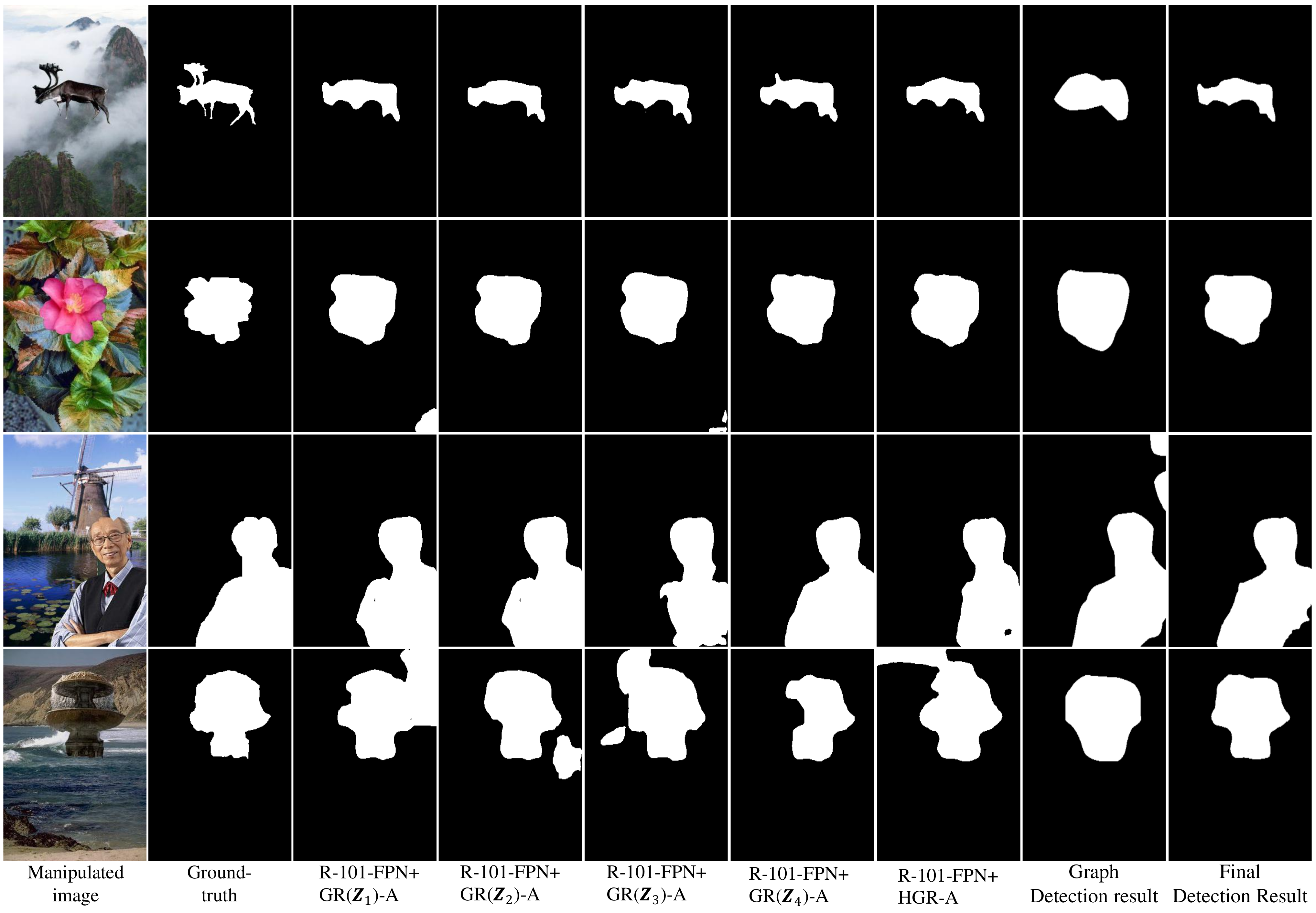}
\caption{Qualitative results on CASIA dataset. The columns from left to right indicate the manipulated images, ground-truth labels, detection results obtained by different settings in Table~\ref{tab:1}. The eighth column illustrates the detection results of the HGRL branch. The last column illustrates the detection results of the HGCN-Net.}
\label{Figure:5}
\end{figure}

{\bf Comparison with State-of-the-arts.} We conduct experiments on the public datasets and compare the performance with prior approaches such as GSR-Net ~\cite{zhou2020generate}, CR-CNN ~\cite{yang2020constrained}, EXIF-Consistency ~\cite{huh2018fighting}. The pixel-level $F_1$ score and MCC are also used for performance evaluation. For the CR-CNN and GSR-Net, we use these networks to train on the training set and evaluate on test sets. The performance of these approaches is shown in Table~\ref{tab:4}. From Table~\ref{tab:4}, it is clear that the proposed HGCN-Net outperforms the CR-CNN and GSR-Net by a large margin on both CASIA and COVER. Moreover, our approach achieves the highest accuracy in Columbia, but slightly lower accuracy in the NIST16. Thus, the proposed HGCN-Net generally outperforms the other approaches. The result demonstrates that feature correlations is beneficial for improving the performance in image manipulation detection tasks.

{\bf Robustness.} In this section, we evaluate the robustness of the proposed HGCN-Net under a variety of common image attacks, including JPEG compression, resizing, Gaussian blur, blur, and rotation. We compare HGCN-Net with GSR-Net and CR-CNN using their publicly available codes. The results on the CASIA and COVER are shown in Figure~\ref{Figure:4}. It can be clearly observed that the proposed HGCN-Net achieves high robustness with regard to all attacks on the CASIA and COVER datasets. This is because the feature correlations between manipulated and non-manipulated regions learned by HGCN-Net are relatively stable under those common attacks.

\begin{figure}[t]
\centering
\includegraphics[width=8.7cm,height=5.1cm]{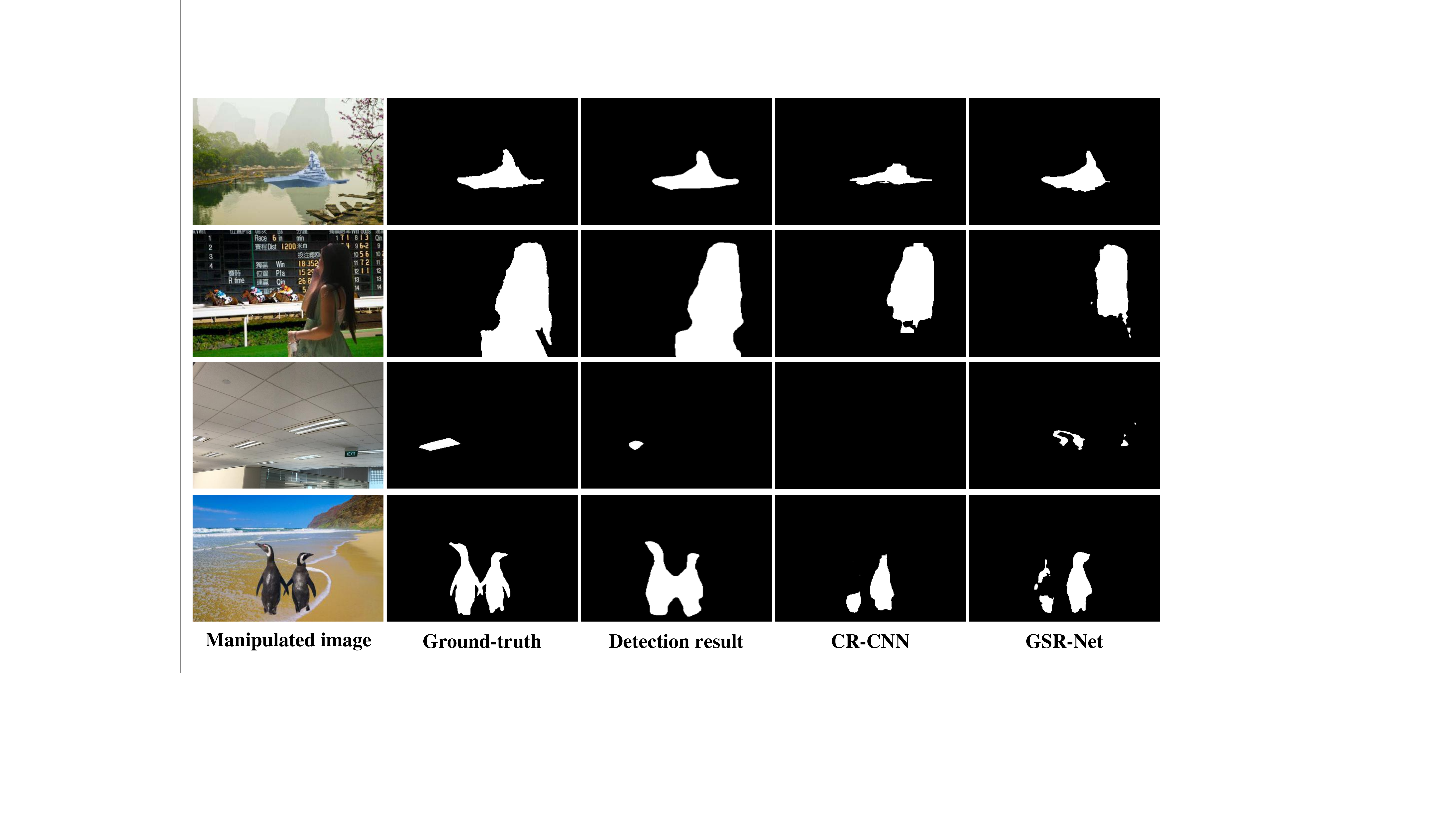}
\caption{Qualitative results of different networks on the public datasets. The first column shows manipulated images on different datasets. The second column illustrates the ground-truth labels. The third column illustrates the final manipulation detection results of the proposed HGCN-Net. The fourth column illustrates the detection results of CR-CNN. The last column illustrates the detection results of GSR-Net.}
\label{Figure:6}
\end{figure}

{\bf Qualitative Results.} We first provide the qualitative result in Figure~\ref{Figure:5} to illustrate the performance of the proposed HGCN-Net with different settings listed in Table~\ref{tab:1}. The results show HGCN-Net with setting R-101-FPN+HGR-C(w/ $\mathcal{L}_S$) achieves the best performance. We also provide qualitative results in Figure~\ref{Figure:6} to visually compare HGCN-Net with CR-CNN and GSR-Net. As shown in Figure 6, the proposed HGCN-Net always produces smoother boundary and more accurate pixel-level localization results for detecting manipulated regions.

\section{Conclusion}

In this paper, we propose the HGCN-Net for image manipulation detection, which performs promising detection accuracy and strong robustness to a variety of common image attacks. This is achieved by sufficiently exploring the feature correlations within the manipulated images. Such feature correlations are captured by the proposed hierarchical graph representation, which is learned from the set of the fully-connected graphs constructed by connecting all grids in the feature maps of a given manipulated image. Experimental results demonstrate the effectiveness and strong robustness of our network on image manipulation detection tasks.

\section*{Acknowledgments}

This work is supported in part by the National Natural Science Foundation of China under Grant 61972205, 62122032, 62032020, in part by the Priority Academic Program Development of Jiangsu Higher Education Institutions (PAPD) fund, and in part by the Collaborative Innovation Center of Atmospheric Environment and Equipment Technology (CICAEET) fund, China.

\appendix
\bibliographystyle{named}
\bibliography{ref}
\end{document}